\documentclass[10pt,twocolumn,letterpaper]{article}

\usepackage[pagenumbers]{cvpr} 

\usepackage{graphicx}
\usepackage{amsmath}
\usepackage{amssymb}
\usepackage{booktabs}

\usepackage[pagebackref,breaklinks,colorlinks]{hyperref}

\usepackage{fontawesome5}
\usepackage{pifont} 

\usepackage[capitalize]{cleveref}
\crefname{section}{Sec.}{Secs.}
\Crefname{section}{Section}{Sections}
\Crefname{table}{Table}{Tables}
\crefname{table}{Tab.}{Tabs.}

\def\cvprPaperID{*****} 
\def\confName{CVPR}
\def\confYear{2022}

\begin{document}

\title{Instance Segmentation under Occlusions via Location-aware Copy-Paste Data Augmentation}

\author{Son Nguyen\textsuperscript{\textdagger}, Mikel Lainsa\textsuperscript{\textdagger}, Hung Dao\textsuperscript{\textdagger}, Daeyoung Kim\textsuperscript{\textdagger}, Giang Nguyen\textsuperscript{\ding{169}}\\
KAIST, South Korea\textsuperscript{\textdagger}, Auburn University, US\textsuperscript{\ding{169}}\\
{\tt\small \{nguyendinhson,mikel,hicehehe,kimd\}@kaist.ac.kr\textsuperscript{\textdagger}, nguyengiangbkhn@gmail.com\textsuperscript{\ding{169}} }}


\maketitle

\begin{abstract}
   Occlusion is a long-standing problem in computer vision, particularly in instance segmentation. 
   ACM MMSports 2023 DeepSportRadar has introduced a dataset that focuses on segmenting human subjects within a basketball context and a specialized evaluation metric for occlusion scenarios. 
   Given the modest size of the dataset and the highly deformable nature of the objects to be segmented, this challenge demands the application of robust data augmentation techniques and wisely-chosen deep learning architectures. 
   Our work (ranked $1^{st}$ in the competition) first proposes a novel data augmentation technique, capable of generating more training samples with wider distribution.
   Then, we adopt a new architecture -- Hybrid Task Cascade (HTC) framework with CBNetV2 as backbone and MaskIoU head to improve segmentation performance. 
   Furthermore, we employ a Stochastic Weight Averaging (SWA) training strategy to improve the model's generalization. 
   As a result, we achieve a remarkable occlusion score (OM) of $0.533$ on the challenge dataset, securing the top-1 position on the leaderboard. Source code is available at this \href{https://github.com/nguyendinhson-kaist/MMSports23-Seg-AutoID}{URL}.
\end{abstract}

\section{Introduction}
\label{sec:intro}


Instance segmentation is a crucial task in computer vision, bridging the gap between object detection and semantic segmentation. 
Its applications span various fields, from autonomous driving \cite{de2017semantic} and medical image analysis \cite{} to sports analytics \cite{saito2022mmsports}. 
Recently, deep learning models like Mask R-CNN \cite{he2017mask} and its derivatives have set new benchmarks in accuracy for many computer vision challenges, including instance segmentation. 
As such, these models have become the gold standard, emphasizing their importance in the discipline. 
Additionally, by adopting Transformer backbones, initially intended for Natural Language Processing tasks \cite{vaswani2017attention}, it is possible to achieve rich and meaningful feature representations from images through attention mechanisms \cite{}.


Despite recent advancements, two main challenges persist in instance segmentation: recognizing objects that change shape due to factors like poses or angles (deformation) and identifying objects that are partially or fully hidden behind others (occlusion).
Such complexities are not unique to instance segmentation but also manifest in related domains like object detection and image classification, as evidenced by several studies \cite{wang2020robust, barbu2019objectnet, nguyen2021effectiveness, taesiri2023imagenet, taesiri2022visual}.

The traditional COCO metrics \cite{lin2014microsoft} rely on IoU (Intersection over Union) values to compare model predictions with ground truths in object detection and segmentation.
However, these metrics could not be sufficient and satisfactory for evaluating models in occlusion scenarios. 
In response to this, the \href{http://mmsports.multimedia-computing.de/mmsports2023/challenge.html}{ACM MMSports 2023} challenge has introduced a novel and specialized evaluation metric known as the Occlusion Metric (OM).
The OM metric for the challenge focuses solely on instances that appear partially-visible due to occlusions, as verified by ground-truth annotations. The rationable is that models adept at handling these complex occluded instances should excel at simpler cases. Specifically, OM emphasizes reconnecting occluded pixels. The metric is calculated as the product of two components: Occluded Instance Recall (OIR), which measures the recall of visually split instances, and Disconnected Pixel Recall (DPR), which assesses the recall of pixels separated from the main structure of those split instances.

The challenge's primary aim is to segment human subjects on a basketball court, with potential overlaps in the masks of players and referees. 
Given a dataset of just 324 training images
\footnote{In the context of this competition, the organizers consider train \& val for training, test for validation (can be used many times) and challenge for the test set (can be used only once). More details at this \href{https://github.com/DeepSportradar/instance-segmentation-challenge}{URL}.}
, the competition's organizer presents one of the most challenging tasks to all participants.

In this work, we propose a novel data augmentation pipeline, and a robust Hybrid Task Cascade or HTC-based model \cite{chen2019hybrid} with a transformer backbone. 
We then conduct experiments to validate our proposed method on the validation set to determine the optimal hyper-parameters. 
Finally, we introduce an effective training strategy to improve the model performance on the final challenge (test) set.

Our contributions are summarized as follows:
\begin{itemize}
    \item First, we introduce a novel location-aware (or context-aware \cite{wang2020robust}) copy-paste data augmentation technique to tackle the data-scarce problem given by the competition.
   
    \item Second, we propose a new variant of Hybrid Task Cascade \cite{chen2019hybrid} network that employs CBSwin-Base as the backbone.
    Our model is data-efficient and powerful for instance segmentation to learn rich visual representations. 
\end{itemize}


\section{Methodology}
\label{sec:methodology}

We divide this section into three parts: data augmentation strategy, model design, and training strategy. 
Sec.~\ref{subsec:data_aug} demonstrates all data augmentation techniques used in this work. 
Then, we present all the details about our segmentation model in Sec.~\ref{subsec:model_design}.
Sec.~\ref{subsec:training_strat} will focus on how we perform experiments on the proposed model.

\subsection{Data augmentation}
\label{subsec:data_aug}
\begin{figure}
    \centering
    \includegraphics[width=1\linewidth]{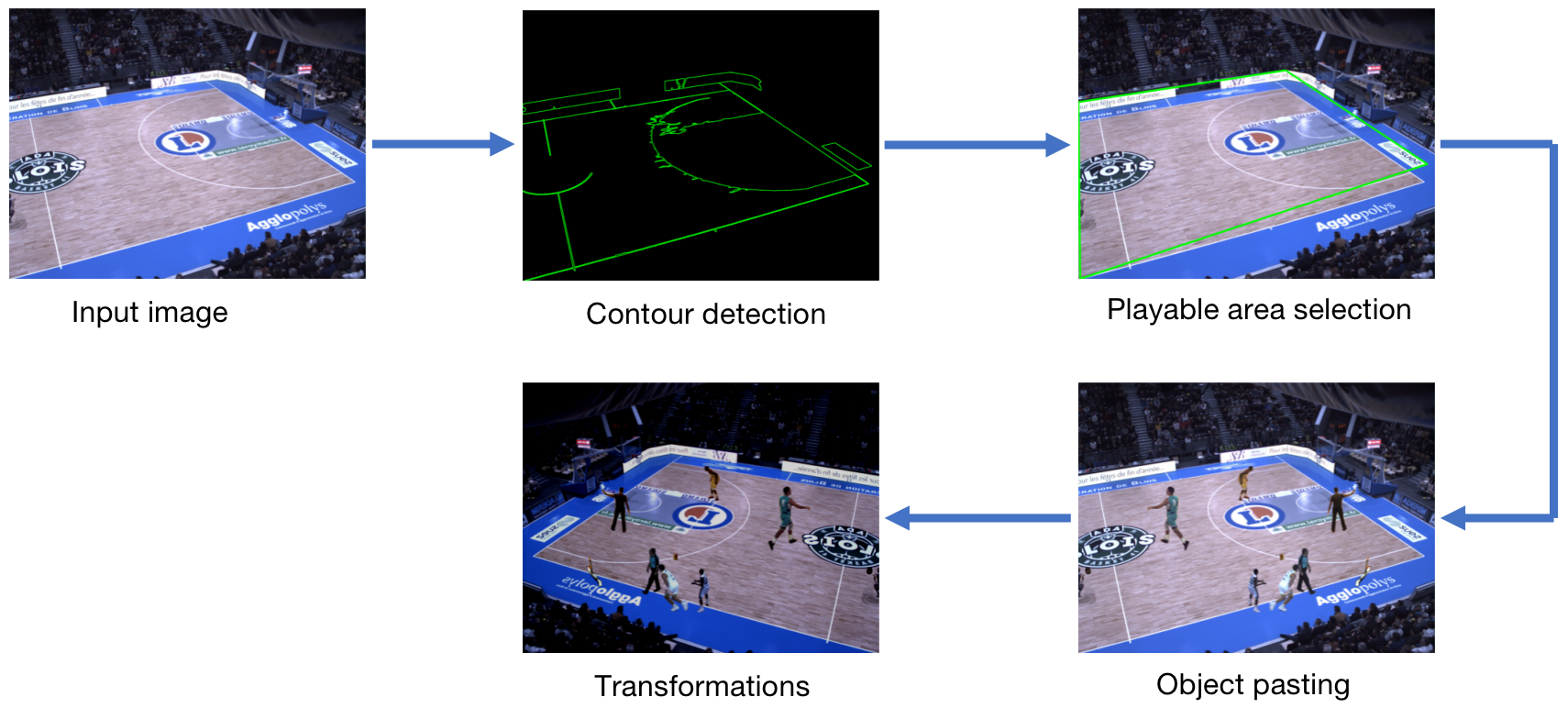}
    \caption{Our location-aware copy-paste data augmentation technique. 
    From an input image of a baseketball court, we perform countour detection, color detection, and Hough line transform to detect the playable area.
    Then, we strategically paste human entities onto the field.
    Finally, photometric distortions and geometric transforms are employed on such human entities.}
    \label{fig:court_estimate}
\end{figure}

Considering the constrained size of the DeepSportRadar instance segmentation dataset \cite{van2022deepsportradar}, leveraging data augmentation is imperative for achieving consistent model performance. 
In light of this, we incorporate a specialized copy-paste augmentation technique alongside traditional geometric and photometric transformations (e.g., translation, rotation, or color jitter).
Fig.~\ref{fig:court_estimate} depicts our copy-paste data augmentation technique.

We refer to ground-truth instances (human objects or basketball players) in the dataset as entities.
To accurately position and paste entities on the court, we build a custom basketball court detector.
This detector leverages traditional computer vision techniques, such as contour detection, color detection, and the Hough line transform (see our \href{https://github.com/nguyendinhson-kaist/MMSports23-Seg-AutoID/blob/master/utils/transforms/special_copypaste.py}{implementation}), to accurately draw the basketball court's boundaries.
Entities are then strategically pasted at random coordinates within the detected playable region.
If the basketball court detector malfunctions or could not precisely identify the boundary, we use default coordinates for entities being pasted as follows:

\begin{equation}
  \frac{w}{5} \leq x_{min} \leq w
  \label{eq:important1}
\end{equation}
\begin{equation}
  0 \leq x_{min} \leq w - \frac{w}{5}
  \label{eq:important2}
\end{equation}
\begin{equation}
  \frac{h}{2} - \frac{h}{5} \leq y_{min} \leq \frac{h}{2} + \frac{h}{5}
  \label{eq:important3}
\end{equation}

where \(w\) and \(h\) denote width and height of the input image, (\(x_{min}\) and \(y_{min}\)) denote the coordinate of an entity being pasted. 
Eq. (3) defines the vertical (y-axis) boundaries of the pasting area, while Eq. (1) and (2) determine the horizontal (x-axis) boundaries. 
We use Eq. (1) for images capturing the left side of the basketball court, and Eq (2) to those showing the right side.

In our novel copy-paste approach, we extract entities from the training samples using their ground-truth segmentation masks. 
Each training image is subjected to an 80\% probability of having additional entities pasted within the basketball court region.
We randomize the number of entities pasted, but it is empirically constrained to a maximum of 40 players.

We apply an occlusion module after the copy-paste augmentation.
That is, once an entity is pasted, this module will have a 70\% chance of overlaying \emph{an additional entity} at the top-left quadrant of the initial entity, thereby simulating realistic occlusion scenarios observed in the training set.


\subsection{Model design}
\label{subsec:model_design}

Hybrid Task Cascade (HTC) \cite{chen2019hybrid} is our primary model architecture.
Instead of relying on the ResNet backbones \cite{he2016deep}, we decide to use the CBSwin-Base backbone \cite{liang2022cbnet} (coupled with a Feature Pyramid Network named CB-FPN) after a careful consideration of the trade-off between model size and available training resources.

\begin{figure*}
    \centering
    \includegraphics[width=0.9\linewidth]{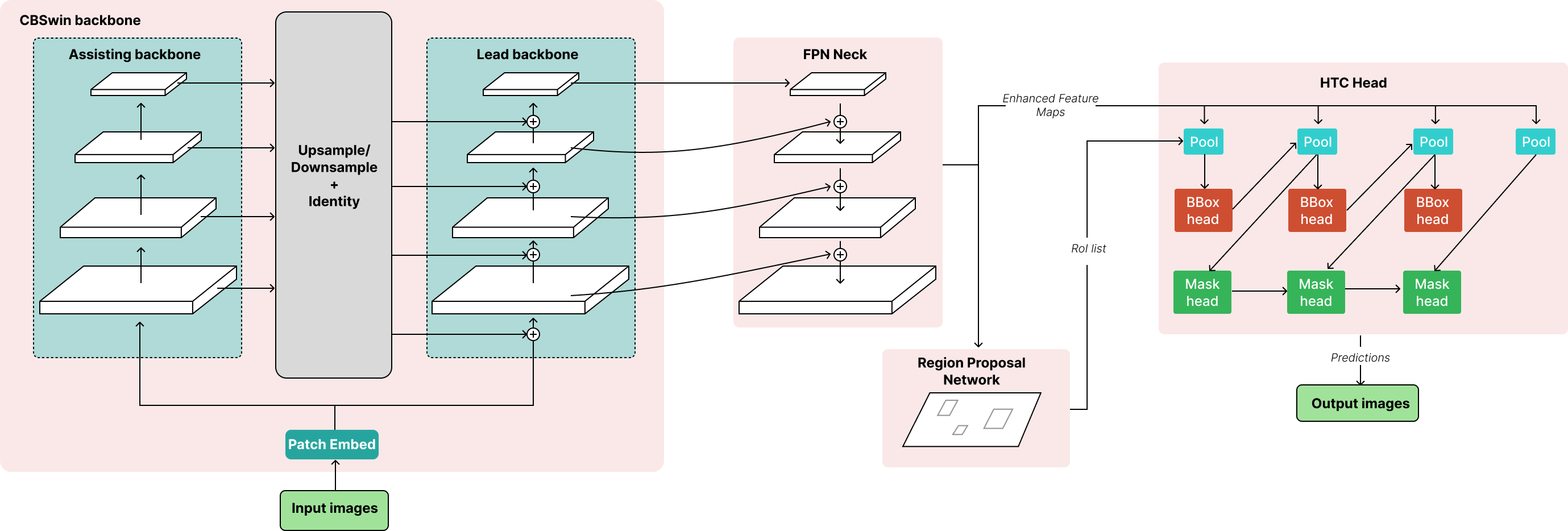}
    \caption{Our proposed instance segmentation architecture. 
    Ours follows the HTC architecture \cite{chen2019hybrid} with some modifications. 
    (1) We replace the main backbone by a CBSwin-Base backbone for better visual representations. 
    (2) We add a MaskIoU head for each mask head of HTC model to improve the quality of mask predictions. 
    (3) We expand the size of the mask features produced by RoIPooling layer from 14x14 to 20x20.}
    \label{fig:architecture}
\end{figure*}

Furthermore, we use MaskIoU \cite{huang2019mask} heads to help the model in learning high-quality mask generation, diverging from the conventional approach of sharing confidence scores with the classification head. 
In the context of occlusion scenarios, the quality of the mask is highly concerned; therefore, we choose to augment the mask predictions as well. 
To achieve this, we expand the mask feature size extracted by the ROIPooling layer from $14\times14$ to $20\times20$. 
This modification not only improves the mask quality but also helps the model to reconstruct object details, even in situations of severe occlusion.
We present the model architecture in Fig.~\ref{fig:architecture}.

\subsection{Training strategy}
\label{subsec:training_strat}

We train and evaluate our model from scratches without any pretrained weights. 
Before training, we need to prepare objects that will be pasted by our copy-paste algorithm. 
As such, we extract all human instances with their corresponding mask annotations and their cropped images from training data, then save them locally.
During augmentation, the copy-paste algorithm will randomly select candidates and define the pasting area for each.
When the copy-paste process is done, we continue the preprocessing with other base transformations: randomly resizing, photometric distortions, and geometric transforms. 
The augmented images are then cropped and padded to a fixed size before being fed into the model.

To tune model hyperparameters, we split 324 training images into 2 sets: 260 images for training and 64 images for validation. After we determine the optimal hyperparameter set, we fix all these values and train the final model on whole training data. 
To improve the model's generalization capability, we also leverage the Stochastic Weight Averaging (SWA) \cite{izmailov2018averaging} training strategy. 
Specifically, after the main training process, we further train the model for 12 more epochs, and save a checkpoint for each epoch.
After the SWA training, we perform average over all 12 saved checkpoints to get the final weights and submit the model.


\section{Experiments}
\label{sec:experiments}

We divide this section into two main parts: the training details and the experimental results.
In Sec.~\ref{subsec:training_details}, we begin by detailing our preprocessing steps and the tools utilized for model development. We then delve into the specifics of our training strategy.
Subsequently, in Sec.~\ref{subsec:exp_results}, we present the outcome of our various training strategies.
\subsection{Training details}
\label{subsec:training_details}
Our models adopt the MMDetection toolbox \cite{chen2019mmdetection} for fast development process. 
All experiments are conducted on two NVIDIA A100-SXM4 GPUs, each equipped with 40GB of memory. 
The copy-paste data augmentation is followed by randomly resizing to one of the following scales: (3680, 3080), (3200, 2400), (2680, 2080), (2000, 1400), (1920, 1440), (1800, 1200), (1600, 1024), (1333, 800), (1624, 1234), (2336, 1752), (2456, 2054). 
A series of photometric distortions, geometric transformations, and cropping operations are subsequently randomly applied on the training data. 
In the final step of preprocessing, all images are consistently resized and padded to a standardized resolution of $1760\times1280$.

For the training process, we use Adam optimizer with decoupled weight decay (AdamW).
The learning rate is set to $1e^{-4}$ and weight decay is configured to $2e^{-2}$ to prevent overfitting. 
When the model starts to converge, we use the SWA training strategy to finetune the model for an additional 12 epochs. Throughout the SWA training process, the optimizer is also AdamW with a constant learning rate of $1e^{-4}$. 
After the entire training process, we take the average of all checkpoints of 12 epochs to get the final model. 
During training, we adjust the loss weight of the mask head, elevating it from $1.0$ to $2.0$. 
This modification aims to emphasize the importance of accurate mask generation within the model's learning process.

\subsection{Experiment results}
\label{subsec:exp_results}

\begin{table*}[hbt!]
    \centering
    \begin{tabular}{c c c c c}
    \hline
        Model & Training set & Test set & Epochs & OM\\
        \hline
        HTC-CBSwinB & train & val & 720 & 0.514\\
        HTC-CBSwinB & train+val & test & 720 & 0.433\\
        HTC-CBSwinB & train+val & test & 1200 & 0.495\\
        HTC-CBSwinB+SWA & train+val & test & 1200+12 & \textbf{0.533}\\
        \hline
    \end{tabular}
    \caption{Experimental results of our model (ranked $1^{st}$) in ACM MMSports 2023 segmentation challenge.}
    \label{tab:ex_result}
\end{table*}

Following the training strategy outlined in Sec.~\ref{subsec:training_strat}, our CBSwin-Base + HTC model achieves an OM (occlusion) score of $0.514$ after 720 epochs of training (see Table.~\ref{tab:ex_result}). 
Based on these results, we keep the same configuration to train the model on more samples (i.e., training + validation set). 
In our initial attempt on the test set, we get an OM score of $0.433$. 
Recognizing the potential for longer training, we decide to resume the training from the point of the last experiment. 
With this extended training time from 720 to 1200 epochs, a simple yet effective idea, we obtain an OM score of $0.495$. 
This shows an impressive improvement of $0.062$ points over the previous attempt. 
Finally, we achieve the final OM score of $0.533$ by leveraging the SWA training strategy \cite{izmailov2018averaging}, concluding our model with the highest score in the MMSports 2023 challenge.

\section{Conclusion}

In this paper, we have reported key methods and techniques used to address the ACM MMSports 2023 Instance Segmentation problem. 
In order to tackle the occlusion in the segmentation task, we leverage a powerful HTC architecture with CBSwin-Base backbone and introduce a novel location-aware copy-paste data augmentation that can be applied arbitrarily on data-scarce segmentation applications.
The experimental results demonstrate the effects of our method to achieve a state-of-the-art result (ranked $1^{st}$ with $0.533$ OM) on the test set without any additional data beyond just 324 given samples or pretraining.

{\small
\bibliographystyle{ieee_fullname}
\bibliography{egbib}

\begin{thebibliography}{10}\itemsep=-1pt

\bibitem{barbu2019objectnet}
Andrei Barbu, David Mayo, Julian Alverio, William Luo, Christopher Wang, Dan Gutfreund, Josh Tenenbaum, and Boris Katz.
\newblock Objectnet: A large-scale bias-controlled dataset for pushing the limits of object recognition models.
\newblock {\em Advances in neural information processing systems}, 32, 2019.

\bibitem{chen2019hybrid}
Kai Chen, Jiangmiao Pang, Jiaqi Wang, Yu Xiong, Xiaoxiao Li, Shuyang Sun, Wansen Feng, Ziwei Liu, Jianping Shi, Wanli Ouyang, et~al.
\newblock Hybrid task cascade for instance segmentation.
\newblock In {\em Proceedings of the IEEE/CVF conference on computer vision and pattern recognition}, pages 4974--4983, 2019.

\bibitem{chen2019mmdetection}
Kai Chen, Jiaqi Wang, Jiangmiao Pang, Yuhang Cao, Yu Xiong, Xiaoxiao Li, Shuyang Sun, Wansen Feng, Ziwei Liu, Jiarui Xu, et~al.
\newblock Mmdetection: Open mmlab detection toolbox and benchmark.
\newblock {\em arXiv preprint arXiv:1906.07155}, 2019.

\bibitem{de2017semantic}
Bert De~Brabandere, Davy Neven, and Luc Van~Gool.
\newblock Semantic instance segmentation for autonomous driving.
\newblock In {\em Proceedings of the IEEE Conference on Computer Vision and Pattern Recognition Workshops}, pages 7--9, 2017.

\bibitem{he2017mask}
Kaiming He, Georgia Gkioxari, Piotr Doll{\'a}r, and Ross Girshick.
\newblock Mask r-cnn.
\newblock In {\em Proceedings of the IEEE international conference on computer vision}, pages 2961--2969, 2017.

\bibitem{he2016deep}
Kaiming He, Xiangyu Zhang, Shaoqing Ren, and Jian Sun.
\newblock Deep residual learning for image recognition.
\newblock In {\em Proceedings of the IEEE conference on computer vision and pattern recognition}, pages 770--778, 2016.

\bibitem{huang2019mask}
Zhaojin Huang, Lichao Huang, Yongchao Gong, Chang Huang, and Xinggang Wang.
\newblock Mask scoring r-cnn.
\newblock In {\em Proceedings of the IEEE/CVF conference on computer vision and pattern recognition}, pages 6409--6418, 2019.

\bibitem{izmailov2018averaging}
P Izmailov, AG Wilson, D Podoprikhin, D Vetrov, and T Garipov.
\newblock Averaging weights leads to wider optima and better generalization.
\newblock In {\em 34th Conference on Uncertainty in Artificial Intelligence 2018, UAI 2018}, pages 876--885, 2018.

\bibitem{liang2022cbnet}
Tingting Liang, Xiaojie Chu, Yudong Liu, Yongtao Wang, Zhi Tang, Wei Chu, Jingdong Chen, and Haibin Ling.
\newblock Cbnet: A composite backbone network architecture for object detection.
\newblock {\em IEEE Transactions on Image Processing}, 31:6893--6906, 2022.

\bibitem{lin2014microsoft}
Tsung-Yi Lin, Michael Maire, Serge Belongie, James Hays, Pietro Perona, Deva Ramanan, Piotr Doll{\'a}r, and C~Lawrence Zitnick.
\newblock Microsoft coco: Common objects in context.
\newblock In {\em Computer Vision--ECCV 2014: 13th European Conference, Zurich, Switzerland, September 6-12, 2014, Proceedings, Part V 13}, pages 740--755. Springer, 2014.

\bibitem{nguyen2021effectiveness}
Giang Nguyen, Daeyoung Kim, and Anh Nguyen.
\newblock The effectiveness of feature attribution methods and its correlation with automatic evaluation scores.
\newblock {\em Advances in Neural Information Processing Systems}, 34:26422--26436, 2021.

\bibitem{saito2022mmsports}
Hideo Saito, Thomas~B Moeslund, and Rainer Lienhart.
\newblock Mmsports' 22: 5th international acm workshop on multimedia content analysis in sports.
\newblock In {\em Proceedings of the 30th ACM International Conference on Multimedia}, pages 7386--7388, 2022.

\bibitem{taesiri2023imagenet}
Mohammad~Reza Taesiri, Giang Nguyen, Sarra Habchi, Cor-Paul Bezemer, and Anh Nguyen.
\newblock Imagenet-hard: The hardest images remaining from a study of the power of zoom and spatial biases in image classification.
\newblock In {\em Thirty-seventh Conference on Neural Information Processing Systems Datasets and Benchmarks Track}, 2023.

\bibitem{taesiri2022visual}
Mohammad~Reza Taesiri, Giang Nguyen, and Anh Nguyen.
\newblock Visual correspondence-based explanations improve ai robustness and human-ai team accuracy.
\newblock {\em Advances in Neural Information Processing Systems}, 35:34287--34301, 2022.

\bibitem{van2022deepsportradar}
Gabriel Van~Zandycke, Vladimir Somers, Maxime Istasse, Carlo~Del Don, and Davide Zambrano.
\newblock Deepsportradar-v1: Computer vision dataset for sports understanding with high quality annotations.
\newblock In {\em Proceedings of the 5th International ACM Workshop on Multimedia Content Analysis in Sports}, pages 1--8, 2022.

\bibitem{vaswani2017attention}
Ashish Vaswani, Noam Shazeer, Niki Parmar, Jakob Uszkoreit, Llion Jones, Aidan~N Gomez, {\L}ukasz Kaiser, and Illia Polosukhin.
\newblock Attention is all you need.
\newblock {\em Advances in neural information processing systems}, 30, 2017.

\bibitem{wang2020robust}
Angtian Wang, Yihong Sun, Adam Kortylewski, and Alan~L Yuille.
\newblock Robust object detection under occlusion with context-aware compositionalnets.
\newblock In {\em Proceedings of the IEEE/CVF Conference on Computer Vision and Pattern Recognition}, pages 12645--12654, 2020.

\end{thebibliography}
}

\end{document}